\documentclass[11pt,letterpaper]{article}
\usepackage[hyperref]{emnlp2018}
\usepackage{times}
\usepackage{latexsym}
\usepackage{url}
\usepackage{amsmath}
\usepackage{amsfonts}
\usepackage{graphicx}
\usepackage{booktabs}

\aclfinalcopy % Uncomment this line for the final submission

\newcommand{\word}[1]{\emph{#1}}

\newcommand{\figref}[1]{Figure~\ref{#1}}

\newcommand{\tabref}[1]{Table~\ref{#1}}
\newcommand{\Tabref}[1]{Table~\ref{#1}}
\newcommand{\secref}[1]{Section~\ref{#1}}

\title{Representing Social Media Users for Sarcasm Detection}

\author{Y. Alex Kolchinski \\
  Stanford University \\
  {\tt kolchinski@stanford.edu} \\\And
  Christopher Potts \\
  Stanford University\\
  {\tt cgpotts@stanford.edu} \\}

\date{}

\begin{document}
\maketitle
\begin{abstract}
We explore two methods for representing authors in the context of textual sarcasm detection: a Bayesian approach that directly represents authors' propensities to be sarcastic, and a dense embedding approach that can learn interactions between the author and the text. Using the SARC dataset of Reddit comments, we show that augmenting a bidirectional RNN with these representations improves performance; the Bayesian approach suffices in homogeneous contexts, whereas the added power of the dense embeddings proves valuable in more diverse ones.
%   This paper presents new state of the art results for sarcasm detection on the Reddit Sarcasm benchmark \cite{khodak2017large}. The model presented also learns meaningful representations of users and communities which enable the contextual detection of sarcasm in otherwise ambiguous statements.
\end{abstract}

\section{Introduction}

Irony and sarcasm\footnote{We use ``sarcasm'' to include both sarcasm and irony, as the two are generally conflated in the literature we review.} are extreme examples of context-dependence in language. Given only the text \word{Great idea!}\ or \word{What a hardship!}, we cannot resolve the speaker's intentions unless we have insight into the circumstances of utterance -- who is speaking, and to whom, and how the content relates to the preceding discourse \citep{clark_1996}. While certain texts are biased in favor of sarcastic  uses \cite{kreuz2007lexical,wallace2014humans}, the non-literal nature of this phenomenon ensures that there is an important role for pragmatic inference \citep{Clark:Gerrig:1984}.

The current paper is an in-depth study of one important aspect of the context dependence of sarcasm: the author. Our guiding hypotheses are that authors vary in their propensity for using sarcasm, that this propensity is influenced by more general facts about the context, and that authors have their own particular ways of indicating sarcasm. These hypotheses are well supported by psycholinguistic research \citep{colston2004gender,gibbs2000irony,Dress-etal:2008}, but our ability to test them at scale has until recently been limited by available annotated corpora. With the release of the Self-Annotated Reddit Corpus (SARC), \citet{khodak2017large} have helped to address this limitation. SARC is large and diverse, and its distribution of users across comments and forums makes it particularly well suited to modeling authors and their relationship to sarcasm.

\begin{figure}[t]
  \includegraphics[width=1\linewidth]{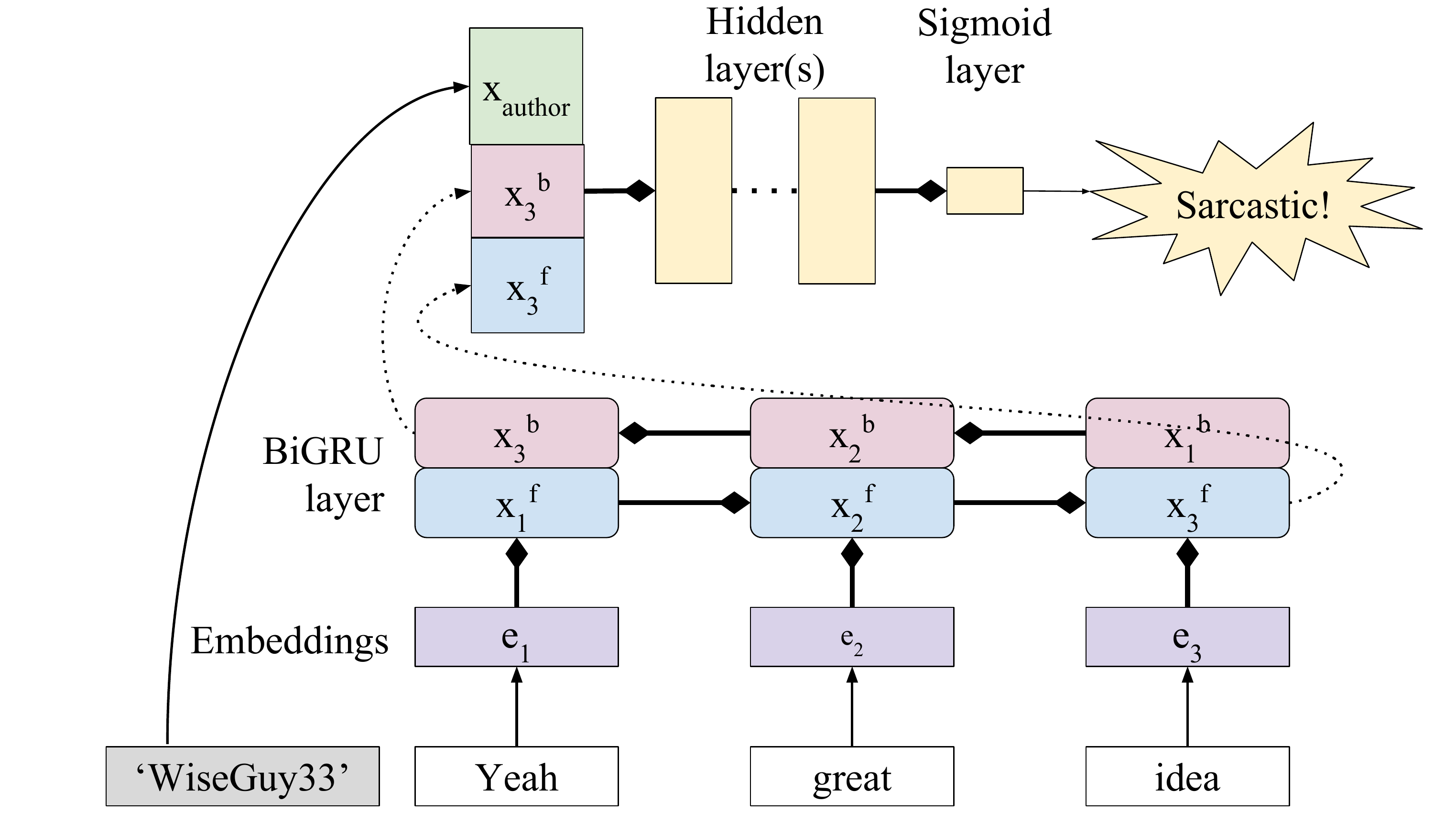}
  \caption{The model architecture. Look-ups are indicated by arrows, dense connections by diamonds. The author embedding can be null (a text-only baseline), a prior reflecting the author's propensity for sarcasm, or a learned embedding. There are potentially multiple layers between the initial example embedding and the output sigmoid layer.}
  \label{fig:model}
\end{figure}

Our core model of comment texts is a bidirectional RNN with GRU cells. To model authors, we propose two strategies for augmenting these RNN representations: a simple Bayesian method that captures only an author's raw propensity for sarcasm, and a dense embedding method that allows for complex interactions between author and text (\figref{fig:model}). We find that, on SARC, the simple Bayesian approach does remarkably well, especially in smaller, more focused forums. On the full SARC dataset, author embeddings are able to encode more kinds of variation and interaction with the text, and thus they achieve the highest predictive accuracy. These findings extend and reinforce the prior work on user-level modeling for sarcasm (\secref{sec:lit}), and they indicate that simple representation methods are effective here.

\section{Previous Work}\label{sec:lit}

A substantial literature exists around sarcasm detection. Many of the prior studies focus on the analysis of Twitter posts, which lend themselves well to sarcasm detection with NLP methods because they are available in large quantities, they tend to correspond roughly to a single utterance, and users' hashtags in tweets (e.g., \#sarcasm, \#not) can provide imperfect but useful labels. A central theme of this literature is that bringing in contextual features helps performance.

\citet{gonzalez2011identifying} trained classifiers using a combination of lexical and pragmatic features, including emoticons and whether the user was responding to another tweet (see also \citealt{felbo2017using}). \citet{bamman2015contextualized} extend this kind of analysis with additional information about the context. Of special interest here are their contextual features: the author's historical sentiment, topics, and terms; the addressee; and features drawn from historical interactions between the author and addressee. The study finds most features to be useful, but a model trained on the tweet and author features alone achieved essentially the same performance (84.9\% accuracy) as a model trained on all features (85.1\%).

In a similar vein, \citet{rajadesingan2015sarcasm} used a complex combination of features from users' Twitter histories, including sentiment, grammar, and word choice, as inputs into their model, and report a $\approx$7\% gain in classification accuracy upon adding these features to a baseline n-gram classifier.

Recent papers have also applied deep learning methods to detecting sarcastic tweets. \citet{poria2016deeper} use a combination convolutional--SVM architecture with auxiliary sentiment input features. The architecture of \citet{zhang2016tweet} includes an RNN, and uses contextual features as well as tweet text for inputs. 

\citet{amir2016modelling} extend the work of \citeauthor{bamman2015contextualized} by generating author embeddings to reflect users' word-usage patterns (but not sarcasm history) in a manner similar to the paragraph vectors introduced by \citet{le2014distributed}. With the inclusion of these embeddings, their convolutional neural network (CNN) achieves a 2\% gain in accuracy over that of \citeauthor{bamman2015contextualized}.

\citet{ghosh2017magnets} present a combination CNN/LSTM (long short-term memory RNN) architecture that takes as inputs user affect inferred from recent tweets as well as the text of the tweet and that of the parent tweet. When a tweet was addressed to someone by name, the name of the addressee was included in the text representation of the tweet, providing a loose link between interlocutors \citep{West-etal:2014} and a $\approx$1\% gain in performance for some data sets.

There has also been a small amount of previous work on Reddit data for sarcasm \citep{tay2018reasoning,ghosh20181}. \citet{wallace2014humans} explore a hand-labeled dataset of $\approx$3K Reddit comments from six subreddits. They report that, when human graders attempted to mark comments as sarcastic or not sarcastic, they needed additional context like subreddit norms and author history roughly 30\% of the time, and that the comments which graders found ambiguous were largely the same as those on which a baseline bag-of-words classifier tended to make mistakes. In a follow-up study, \citet{wallace2015sparse} find that semantic cues for sarcasm differ by subreddit, and they show classifier accuracy gains when modeling subreddit-specific variation.

The work that is closest to our own is that of \citet{Hazarika-etal:2018}, who also experiment on the SARC dataset. Their model learns author, forum, and text embeddings, and they show that all three kinds of representation contribute positively to the overall performance. We take a much simpler approach to author embeddings and do not include forum embeddings, and we report comparable performance (\secref{sec:results}). We take this as further indication of the value of author features for modeling sarcasm.

\section{The SARC Dataset}\label{sec:dataset}

The Self-Annotated Reddit Corpus (SARC) was created by \citet{khodak2017large}.\footnote{\url{http://nlp.cs.princeton.edu/SARC/2.0/}} It includes an unprecedented 533M comments. The corpus is self-annotated in the sense that a comment is considered sarcastic if its author marked it with the ``/s'' tag. As a result, the positive examples are essentially those which the authors considered ambiguous enough to explicitly tag as sarcastic, meaning that the prediction problem is actually to identify which comments are not only sarcastic but both sarcastic and not obviously so. 

The dataset is filtered in numerous ways, and has good precision (only $\approx$1\% false positive rate) but poor recall (2\% false negatives relative to 0.25\% true positives, or $\approx$11\% recall). To alleviate the issues caused by low recall, the dataset also includes a balanced sample, where comments are supplied in pairs, both responding to the same parent comment and with exactly one of the two tagged as sarcastic. All comments are accompanied with ancestor comments from the original conversation, author information, and a score as voted on by Reddit users.

This dataset presents numerous advantages for sarcasm detection. For one, it is vastly larger than past sarcasm datasets, which enables the training of more sophisticated models. In addition, most work in sarcasm detection has focused on tweets, which are very short and tend to use abbreviated and atypical language. Reddit comments are not constrained by length and are therefore more representative of how people typically write. Finally, Reddit is organized into topically-defined communities known as subreddits, each of which has its own community norms and linguistic patterns. By making available large amounts of data from a number of subreddits, SARC facilitates the comparative analysis of subreddits, and more generally provides a view into the differences between communities.

\Tabref{tab:dataset} provides basic statistics on the entire corpus as well as the subreddits that we focus on in our experiments.

\begin{table}[t]
\centering
\setlength{\tabcolsep}{4pt}
\begin{tabular}{l r r}
\toprule
  & Comments & \% sarcastic \\
\midrule
Entire corpus (bal.) & 257,082 & 50.00\\[1ex]
%Entire corpus (unbal.) & & \\[1ex]
r/politics (bal.) & 13,668 & 50.00 \\
r/politics (unbal.) & 309,925 & 3.06\\[1ex]
r/AskReddit (bal.) & 11,660 & 50.00\\
r/AskReddit (unbal) & 1,548,803 & 0.53 \\
\bottomrule
\end{tabular}
\caption{Basic statistics for SARC.}
\label{tab:dataset}
\end{table}

% Explain more about subreddits, balanced/unbalanced datasets
% Include table here - full dataset balanced/unbalanced, r/politics balanced/unbalanced, r/AskReddit balanced/unbalanced. For each: number of total comments/comment pairs, and proportion that are sarcastic or not sarcastic.

\section{Models}\label{sec:models}

Our baseline model is a bidirectional RNN with GRU cells (BiGRU; \citealt{D14-1179}). We tried variants with LSTM cells and did not observe a significant difference in performance. We therefore chose to use GRU cells as the model with fewer parameters.\footnote{Our models and associated experiment code are available at \url{https://github.com/kolchinski/reddit-sarc}}

The inputs to the BiGRU model are users' comments, which are split into words (and in the case of conjunctions, subwords) and punctuation marks and are converted to word vectors. The final states of the two directions of the BiGRU are concatenated with each other and run through either a single fully-connected linear layer or two fully-connected linear layers with a rectified linear unit in between. The output of the final linear layer is fed through a sigmoid function which outputs the estimated probability of sarcasm. This baseline does not take author information into account: for each comment, only the words of the comment are considered as inputs.

The \emph{Bayesian prior} model extends the BiGRU with the sarcastic and non-sarcastic comment counts for authors seen in the training data, which serves as a prior for sarcasm frequency. This version of the model takes as inputs both a representation of the comment and the author representation $x_{\text{author}}\in \mathbb{Z}^2_{\geq0}$ to estimate the probability of sarcasm. The model can be interpreted as computing a posterior probability of sarcasm given both the comment and the prior of previous sarcastic and non-sarcastic comment counts -- author modeling reduced to a Bernoulli prior. For previously unseen authors, $x_{\text{author}}$ is set to $(0,0)$.

\newcommand{\xUNK}{x_{\textsc{unk}}}

The \emph{author embedding} approach extends the baseline BiGRU in a more sophisticated way. Here, each author seen in the training data is associated with a randomly initialized embedding vector $x_{\text{author}} \in \mathbb{R}^{15}$, which is then provided as an input to the model along with a representation of the words of the comment. A special randomly initialized vector $\xUNK$ is used for previously unseen authors. The author embeddings are updated during training, with the goal of learning more sophisticated individualized patterns of sarcasm than the  Bayesian prior allows. We experimented with training the $\xUNK$ vector on infrequently-seen authors (fewer than 5 comments in the training set) instead of using a random vector, and found some suggestions of improved performance. However, as the differences in performance were not substantial enough to change the relative performance of the different models, we report the results for the simpler random-$\xUNK$ model.

\section{Experiments}

\newcommand{\result}[3]{#1 {\footnotesize[#2, #3]}}

\newcommand{\colcenter}[1]{\multicolumn{1}{c}{#1}}

\begin{table*}[t!]
\centering

\renewcommand{\textbf}[1]{#1}
\renewcommand{\emph}[1]{#1}

\resizebox{\textwidth}{!}{%
\begin{tabular}{l *{5}{l} }
\toprule
               &                & \multicolumn{2}{c}{r/politics}   & \multicolumn{2}{c}{r/AskReddit}  \\
               & \colcenter{Full balanced}  & \colcenter{balanced} & \colcenter{unbalanced}            & \colcenter{balanced} & \colcenter{unbalanced} \\
\midrule               
No embed                & \result{\emph{74.8}}{74.6}{74.9}   & \result{74.3}{74.1}{74.6}           & \result{58.7}{58.2}{59.1}          & \result{64.3}{63.4}{65.2}           & \result{56.9}{56.6}{57.2}   \\
Bayesian prior          & \result{74.0}{73.7}{74.3}          & \result{\textbf{77.6}}{77.4}{77.9}  & \result{\textbf{64.7}}{64.6}{64.8} & \result{\textbf{69.1}}{68.8}{69.4}  & \result{\textbf{57.7}}{57.6}{57.7}   \\
15d embed               & \result{\textbf{75.3}}{74.8}{75.7} & \result{75.1}{74.4}{75.8}          & \result{\emph{62.0}}{59.9}{63.9}   & \result{66.0}{65.1}{66.8}            & \result{\emph{57.1}}{56.6}{57.6}   \\[1ex]
\citet{khodak2017large} & 75.8                             & 76.5                              & 27.0   & -- & -- \\
CASCADE                 & 77.0                             & 75.0                              & --     & -- & -- \\
\bottomrule
\end{tabular}
}
\caption{Mean macro-averaged F1 scores with bootstrapped 95\% confidence intervals, based on 10 runs. 
%Scores in italics indicate systems for which $p \geq 0.05$ in a comparison with the best system (bold), according to a Wilcoxon signed-rank test.
CASCADE is the best system of \citet{Hazarika-etal:2018}, and we report the strongest baseline numbers established by  \citet{khodak2017large}.}
\label{results-table}
\end{table*}

We conducted three sets of experiments, one for each model, to evaluate the effectiveness of the different approaches to author modeling. Each set of experiments was conducted on five datasets: the balanced version of the entire corpus as well as the balanced and unbalanced versions of the r/politics and r/AskReddit subcorpora (\tabref{tab:dataset}).

In all cases, the raw comment data was tokenized into words and punctuation marks, with components of contractions treated as individual words.  We mapped tokens to FastText embedding vectors which had been trained, using subword infomation, on Wikipedia 2017, the UMBC webbase corpus, and the statmt.org news dataset \cite{mikolov2018advances}. While vectors existed for nearly 100\% of tokens generated, exceptions were mapped to a randomly initialized UNK vector.

All models were trained with early stopping on a randomly partitioned holdout set of either 5\% of the data for balanced subreddit corpora or 1\% for the others. The performance of the model, as used for hyperparameter tuning, was evaluated against a second holdout set, generated in the same manner as the first holdout set but disjoint from both it and the portion of the data used for training. 

Hyperparameters were tuned to maximize model performance as evaluated in this manner, starting with a randomized search process and fine-tuned manually.  The final evaluation was conducted against the test set, with a single randomly partitioned holdout set from the training data again used for early stopping.
We applied dropout \citep{srivastava2014dropout} during training before and between all linear layers. For additional regularization, we also applied an l2-norm penalty to the linear weights but not to the GRU weights.

We attempted other model variations, including multiple GRU layers and an attention mechanism for GRU outputs, but did not observe any gains in performance from the larger models.

\section{Results and Discussion}\label{sec:results}

\begin{table*}[ht]
\centering
\setlength{\tabcolsep}{4pt}
\begin{tabular}{r@{ \ }p{6cm} c c c c}
\toprule
                              & & & \multicolumn{3}{c}{Model Predictions of p(sarcastic)}  \\
 & Reddit comment & Sarcastic? & No user rep. & Bayesian & Multidimensional  \\
\midrule
1. & Good thing Trump is going to bring back all those low education high paying jobs.
 & Yes & .45 & .68 & .84\\ \midrule
2. & lol woops!
 & No & .78 & .36 & .25\\ \midrule
3. & The most ubiquitous undergarments I see these days. 
 & Yes & .15 & .17 & .79\\ \midrule
4. & Such a deep confession, and it doesn't sound like the guy who wrote it is an asshole at all. 
 & Yes & .33 & .45 & .86\\ \midrule
5. & It's not entirely impossible that there are recipe's that have yet to be discovered.
 & No & .23 & .23 & .81\\
\bottomrule
\end{tabular}
\caption{Examples selected to highlight differences between the models.}
\label{tab:examples}
\end{table*}

\subsection{Quantitive assessment}

\Tabref{results-table} reports the means of 10 runs to control for variation deriving from randomness in the optimization process \citep{reimers2017reporting}.

Where there is overlap between our experiments and those of \citet{Hazarika-etal:2018} (CASCADE), our model is highly competitive. We slightly under-perform on the full balanced dataset but come out ahead on r/politics. This is striking because our model makes use of much less information. First, unlike CASCADE, we do not have forum embeddings. Second, CASCADE author embeddings involve  extensive feature engineering including ``stylometric'' and ``personality'' features. Our author embeddings, on the other hand, are either simple empirical estimates (Bayesian priors) or learned embeddings with random initializations, in both cases allowing simpler model specification and training, and more flexibility on the task for which they are used. 

There is also evidence that the BiGRU yields better representations of texts than does \citeauthor{Hazarika-etal:2018}'s CNN-based model. Our `No embed' model is akin to their CASCADE with no contextual features, which achieves only 0.66 on the full balanced corpus and 0.70 on the r/politics balanced dataset. Both numbers are well behind our `No embed'. Unfortunately, we do not have space for a fuller study of the similarities and differences between our model and CASCADE.

Both of our methods for  representing authors perform well. This is perhaps especially striking for the unbalanced experiments, where the percentage of sarcastic comments is tiny (\tabref{tab:dataset}). The two methods perform differently on individual forums than on the full dataset. For the r/politics and r/AskReddit communities, the Bayesian priors give the best results. The situation is reversed for the full dataset, where the high-dimensional embeddings outperform the Bayesian priors. This likely reflects two interacting factors. First, with smaller, more focused forums, it is harder to learn good author embeddings, so the simple prior is more reliable. Second, on the full dataset, there are more examples, and also more complex interactions between authors and their texts, so the added representational power of the embeddings proves justified. 

\subsection{Qualitative comparisons}

\Tabref{tab:examples} provides example predictions from the different models. Each example is taken from the holdout set of a run in which all three models were trained on the same training set and evaluation was conducted on the same holdout set.

For both sarcastic and non-sarcastic comments, author modeling can be helpful for disambiguation. For instance, in examples 1 and 2, omitting author modeling led to incorrect predictions, but including the frequency of the author's sarcasm use alone was enough to change the prediction from incorrect to correct. 

In cases like examples 3 and 4, where the Bayesian prior was insufficient, including a model of the author's individualized patterns of sarcasm was much more powerful. That said, the more complex embedding model can misfire, as in example 5, where the simpler models make a correct prediction but it does not. This appeared to happen more for non-sarcastic examples, where the embedding model would occasionally strongly influence the predicted probability of sarcasm upward. Evidently, authors have more individualized patterns of sarcasm than of non-sarcasm.

Judging by the relative performance of the Bayesian and multidimensional-embedding models (\Tabref{results-table}), the multidimensional model wins more disagreements than it loses with the Bayesian model when there is more training data available. However, when there is not, it overfits to such a degree that its predictions of authors' sarcasm patterns are less useful than the Bayesian approach. This suggests a future direction of exploration: the most useful model of all may be one that expands in complexity for authors with more examples available, and shrinks for those who have fewer.

\section{Conclusion}\label{sec:conclusion}

This paper evaluated two data-driven methods for modeling the role of the author in sarcasm detection. Both prove effective. As shown by \citet{Hazarika-etal:2018}, similar techniques can be extended to other aspects of the context. While our experiments did not support adding these representations, we think listeners rely on them as well, so additional computational modeling work here is likely to prove fruitful.

\bibliography{emnlp2018_kolchinski}
\bibliographystyle{acl_natbib_nourl}

% \appendix

% \section{Supplemental Material}
% \label{sec:supplemental}
% TODO: Edit this out for the paper submission
% The source code for all of the experiments in this paper is at \url{https://github.com/kolchinski/reddit-sarc}; it is open-source under the MIT License.

% \subsection{Preprocessing}
% TODO
% Processing, tokenization decisions, which embeddings we used...

% \subsection{Architectures}
% TODO?
% How exactly were the nets laid out?

% \subsection{Hyperparameters}
% TODO
% Which hyperparameter values did we use?

\end{document}